\documentclass{ws-ijmpd}

\usepackage{hyperref}
\usepackage[super,compress]{cite}
\usepackage{amsmath,amssymb,graphicx}

\newcommand{\aj}{AJ} 
\newcommand{\apjl}{ApJL} 
\newcommand{\apjs}{ApJS} 
\newcommand{\mnras}{MNRAS} 
\newcommand{\nat}{Nature} 

\begin{document}

\markboth{Yu Wang}
{Reflections on ``Can AI Understand Our Universe?''}

%
\catchline{}{}{}{}{}
%

\title{Reflections on ``Can AI Understand Our Universe?''
}

\author{Yu Wang
}

\address{ICRA and Dipartimento di Fisica, Sapienza Universit\`a di Roma, P.le Aldo Moro 5, 00185 Rome, Italy.\\
ICRANet, P.zza della Repubblica 10, 65122 Pescara, Italy. \\
INAF -- Osservatorio Astronomico d'Abruzzo, Via M. Maggini snc, I-64100, Teramo, Italy. \\
wang@icra.it}



\maketitle


\begin{abstract}
This article briefly discusses the philosophical and technical aspects of AI. It focuses on two concepts of understanding: intuition and causality, and highlights three AI technologies: Transformers, chain-of-thought reasoning, and multimodal processing. We anticipate that in principle AI could form understanding, with these technologies representing promising advancements.
\end{abstract}


\ccode{PACS numbers:}


\section{Introduction}

In the article ``Can AI Understand Our Universe?'' \cite{2024arXiv240410019W}, we explore the potential of integrating multi-domain data through a single large language model (LLM) to assess whether AI can function as a unified mind for scientific research. Traditionally, research in astronomy and physics often relies on specialized models tailored to specific tasks and datasets. For instance, estimating redshifts of quasars, classifying astrophysical phenomena, and inferring black hole parameters usually require separate models designed to handle the unique characteristics of each dataset. Our study demonstrates that by fine-tuning a single GPT model, it is possible to handle diverse data and tasks simultaneously.

In our astrophysical classification experiments, we used spectral data from the Sloan Digital Sky Survey (SDSS) \cite{2000AJ....120.1579Y,2023ApJS..267...44A}. The fine-tuned GPT model achieved a classification accuracy of 82\% on spectral data, demonstrating strong capabilities in celestial object recognition. Furthermore, by fine-tuning on quasar spectral data, the model achieved a relative accuracy of 90.66\% in estimating the redshift, indicating that the GPT model is capable for both classification and regression of astrophysical tasks.

For gamma-ray burst (GRB) classification \cite{1993ApJ...413L.101K}, we used spectral information instead of the traditional GRB duration \cite{2016ApJS..223...28N}. The spectral classification results show 95.15\% agreement with those based on GRB durations, revealing distinct spectral properties associated with different origins of GRBs. This offers an alternative approach for classifying GRBs especially for those with ambiguous duration-based classification.

In black hole parameter inference, we simulated $K_{\alpha}$ line data under different spin parameters and viewing angles \cite{2001MNRAS.328L..27W,2009Natur.459..540F}, and fine-tuned the model on these data. Experimental results showed that our model achieved 100\% accuracy in spin direction inference, and relative accuracies of 86.66\% and 94.55\% for spin parameters and viewing angles, respectively, highlighting the potential applications for parameter estimation.

We anticipated that with the exponential growth of observational data from future scientific instruments,  AI will become a central tool for processing large-scale, multi-disciplinary data. The design of future large-scale scientific facilities should consider the requirements from AI which is pushing scientific exploration towards an era of unified and comprehensive data integration.

As we wrote this previous article, we were constantly reflecting: What does understanding mean, and can machines possess the ability to understand? We also observe that currently many studies testing whether LLMs have logical reasoning and inference capabilities, with both positive and negative results \cite{allen2023physics,balepur2024reverse,kambhampati2024can,xie2024memorization,ye2024physics,mirzadeh2024gsm}. However, while our article and these studies focus on testing AI within the context of current technology, we are more interested in, in principle, whether AI can achieve understanding. Specifically, is ``deducing the generative principles of data through data analysis'' achievable. This is not merely a technical question but also philosophical. It involves the definition of understanding, the essence of machine learning, and the differences between human cognition and machine computation. In this article, we will briefly discuss two concepts of understanding in conjunction with current AI technologies.

\section{Two Concepts of Understanding}

In the four-dimensional universe we live, humans perceive space and time through their senses, forming an intuitive understanding of the world. Intuition is not only an essential tool to explore the understanding, but also the foundation in daily life for quick decision-making. With the rise of artificial intelligence, an intriguing question emerges: Can AI construct its own intuition without human-like senses, perhaps even surpassing human perception of the world? Furthermore, can AI, through its unique data-processing capabilities, deeply understand causal relationships and provide new perspectives in fields unexplored by humans?

\subsection{Intuition}

Humans perceive space through binocular vision and time through changes in events and the accumulation of memory \cite{rovelli2019order}. How, then, is this perception formed? By contrast, how does AI perceive space and time? Can AI experience these dimensions differently and develop an intuition beyond human capabilities, enabling it to understand the world from entirely new perspectives?

\subsubsection{Humans Intuition}

Intuition refers to the immediate understanding humans achieve through sensory experiences, forming a "clear image" or "immediate comprehension" \cite{volz2006neuroscience}. It arises from integrating sensory inputs like vision, hearing, and touch, enabling direct perception of the world \cite{merleau2013phenomenology}. For example, when we see a tree, our vision immediately conveys its shape, color, and distance, without requiring reasoning or abstraction. This direct perception facilitates rapid acquisition of information, resulting in an ``evident'' understanding.

Intuition is closely tied to the structure of the human brain \cite{kuo2009intuition, laughlin2016nature}. with sensory systems evolved to process inputs for rapid perception and response. Intuitive experiences aid humans in forming concepts of space, time, and causality. However, these concepts may vary among species. For instance, a vision-deprived organism, though lacking the visual intuition familiar to us, may develop alternative perceptual methods, such as relying on touch or sound, to construct an understanding of the world \cite{finney2003visual, azevedo2024sufficiency}. Such differences in perceptual mechanisms may still support effective understanding, even if the resulting ``intuition'' differs from ours.

Human senses also have significant limitations. Our vision is confined to a narrow band of visible light, and our hearing can only perceive sounds within a limited frequency range. These constraints restrict our ``intuitive'' experience of the world. However, technological advancements allow us to overcome these limitations. Telescopes reveal celestial objects and microscopes expose microscopic organisms beyond the reach of the naked eye\cite{gardner2006james,nan2011five,aasi2015advanced,hu2017taiji}. 

As a result, human understanding of the world is not solely dependent on intuitive experiences but also relies heavily on data-driven reasoning and model construction. Many scientific discoveries and  theoretical concepts, such as quantum mechanics and relativity, lie beyond human sensory perception. These advances rely on data analysis, logical reasoning, and mathematical modeling. In this regard, AI resembles humans in some ways: it lacks human sensory organs but can understand the world through data learning and statistical pattern recognition. AI builds knowledge using multimodal data—numerical values, images, and text—similar to how humans use instruments to observe phenomena beyond their natural sensory capabilities. Its advantage lies in processing volumes and complexities of data beyond human capacity, identifying correlations and causality through mathematical and statistical models. For example, in medical imaging, AI can detect subtle features imperceptible to humans, improving diagnostic accuracy \cite{erickson2017machine,varoquaux2022machine}. Similarly, AI can analyze environmental data to understand complex systems, even without relying on humans ``intuitive experiences'' \cite{hino2018machine, karaca2022multi}.



\subsubsection{AI’s Super-Intuition}

The core function of intuition in the understanding process is to provide a ``mediating framework'' that facilitates quick comprehension without relying on complex reasoning or fundamental theories \cite{kahneman2011thinking, bonnefon2020machine, booch2021thinking}. Intuition serves as a bridge between knowledge and experience, allowing us to rapidly identify key aspects and make conclusions when encountering new information.


AI can simulate the perception of three-dimensional space using 3D video and image data. For example, AI can generate three-dimensional point clouds through stereo vision (using dual cameras) or lidar to perceive object depth and distance. Additionally \cite{li2020deep}, AI can learn spatial object features from multi-angle images in training data, building ``spatial awareness'' through image segmentation and feature detection \cite{remondino2006image}. AI understands time primarily through temporal sequences and rates of change. It captures dynamic variations in events using time-series data (such as sensor readings or video frames). AI ``perceives'' time by analyzing frequencies and sequences of events, computing statistical patterns, and modeling time changes.

In certain aspects, AI can develop ``intuition'' that humans lack, as it processes broader and more detailed data types. AI can also extend perception into dimensions inaccessible to human senses, such as the following examples:

AI's perceptual capabilities can be likened to having hundreds of ``eyes'' (``eyes'' here represent sensory modalities), capable of simultaneously receiving and processing multidimensional data. For example, in neuroscience and genomics research, AI can handle information across thousands of dimensions \cite{vogt2018machine, whalen2022navigating}. For humans, such high-dimensional data is difficult to intuitively comprehend. However, AI can form an ``intuition'' about data structures in high-dimensional spaces, directly identifying correlations across multiple dimensions.

AI possesses ``eyes'' of different scales, enabling it to process data from microscopic to macroscopic levels simultaneously. AI, with the data input from instruments, can observe biological processes at the molecular scale while simultaneously processing astronomical data at galactic scales \cite{jones2019setting, greener2022guide}. This ability allows AI to model and understand spatial phenomena across scales, achieving cross-scale spatial perception \cite{ivezic2020statistics, buchner2024set}.

AI's ``eyes'' that can simultaneously observe the past and the present. While human time perception is linear, AI can perform nonlinear analysis based on data rates of change, and it can even ``leap'' through time to some extent. For instance, in video analysis, AI can skip frames to identify overall trends without parsing each one \cite{zhao2020event,oprea2020review}. This ``nonlinear'' temporal analysis could be applied to prediction and retrospection in the future, enabling AI to effectively analyze the correlation of events at different time segments.

AI also extends beyond the five human senses, expanding the range of human perception. AI can ``see'' the world through photons (vision), ``hear'' the world through the vibration of air molecules (hearing), and ``sense'' the world through other mediators like neutrinos and gravitational waves \cite{aasi2015advanced,an2016neutrino,hu2017taiji}. For instance, AI can directly perceive physical phenomena within stars or supernova explosions using neutrino data, effectively ``seeing'' extreme events in the depths of the universe. AI can also utilize gravitational wave signals to ``hear'' spacetime vibrations, perceiving the motion and evolution of celestial bodies.

AI's ``super-intuition'' makes it powerful for exploring unknown domains, surpassing the capabilities of the five human senses. This multidimensional, multiscale, and multimodal perceptual ability enables AI to identify underlying ``commonalities'' and generate insights in novel ways, achieving deeper comprehension of events and phenomena.

\subsection{Causality}

What does it mean for humans to ``understand the causality of an event''? For example, ``The sky rains, so the ground gets wet'' is a causal relationship. But why does it rain? How does rain cause the ground to become wet? If someone only knows ``The sky rains, so the ground gets wet,'' does that mean they truly understand causality, or are they simply performing pattern matching based on experience? We can continue to ask deeper questions until we reach the fundamental root of the issue: what is the ultimate nature of causality?

\subsubsection{Understanding Causality}

``Understanding the causality of events'' is a complex issue spanning fields like cognitive science, philosophy, and scientific theory \cite{pauli1936space,ellis2005physics,hedstrom2010causal,brukner2014quantum,sloman2015causality}. For humans, causal understanding ranges from simple pattern matching to deep theoretical explanations.

At the most basic level, humans derive causal understanding from observing repeated phenomena. For instance, a child might infer ``Rain causes the ground to get wet'' after repeatedly observing wet ground following rainfall. This causal reasoning is based on correlation and falls under pattern recognition. Though shallow, it is highly efficient in daily life. Human brains excel at quickly grasping causality through intuitive reasoning. For instance, observing a tree branch fall after a strong gust of wind, we intuitively infer that the wind caused it. This type of causal understanding relies on cognitive models in the brain, offering quick but simplified explanations without exploring underlying mechanisms.

Scientific causal understanding is built on theoretical frameworks. For example, the explanation for rain involves air cooling, causing water vapor to condense into droplets, a reasoning rooted in meteorology and physics. Theoretical causality requires extensive background knowledge and abstract thinking, representing a deeper level of understanding. Humans also use counterfactual reasoning to validate causal relationships. For instance, ``Would the ground stay dry if it did not rain?'' This type of reasoning helps distinguish genuine causality from mere correlation by exploring multiple possibilities.

Causal relationships can be investigated layer by layer, but their ultimate boundary depends on several factors. First, our understanding of causality is limited by current scientific knowledge. While we can explain rain formation, further inquiries might lead to global weather systems, Earth’s climate models, or even the sun’s energy transfer. These chains may eventually be traced to fundamental physical laws, but the ultimate explanation for these laws remains a mystery. Ultimately, the inquiry into causality may touch on philosophical questions: What is a ``cause''? Is causality an intrinsic feature of reality, or merely a construct of human cognition? Second, human cognition limits the length of causal chains we can process. Overly complex causal networks may exceed our reasoning capacity, forcing us to rely on simplified models or tools.

\subsubsection{Causality by AI}

AI shares many similarities with humans in its approach to discovering causal relationships, particularly in pattern recognition and hypothesis testing \cite{watanabe1985pattern, rugg2003human}. Humans identify causal patterns through sensory experiences and repeated observations. Similarly, AI extracts causal relationships from complex environments through training on large datasets and statistical pattern recognition, such as analyzing the relationship between weather data and humidity. This experience- and observation-based exploration of causality enables both AI and humans to identify correlations, forming the foundation for deeper causal reasoning.

In both humans and AI, the process of forming and testing hypotheses is another point of similarity. Humans often propose hypotheses intuitively and validate them through experiments or observations, such as hypothesizing, ``If it does not rain, the ground will not get wet,'' and testing this in real-world scenarios. AI, on the other hand, employs causal inference algorithms to automatically generate hypotheses. For instance, it uses causal graphs to test intervention effects between variables, simulates counterfactual scenarios, and verifies causal pathways \cite{zhu2019causal,guo2020survey}. This iterative process of hypothesis and validation allows both humans and AI to refine their causal understanding dynamically.

Additionally, humans and AI share similarities in the fundamental logic of exploring causality through variable integration and hierarchical reasoning. When dealing with complex systems involving multiple variables, humans rely on intuition and experience to identify key variables and construct causal sequences, such as reasoning that ``Clouds form rain, and rain wets the ground.'' AI, however, utilizes parallel computing and causal modeling techniques to extract key variables and identify hierarchical relationships within large datasets, thus unraveling intricate causal networks \cite{wang2023scientific,vallverdu2024causality}. 

\subsubsection{Beyond Human Causality}

Can AI surpass humans in understanding causality? The answer depends on how 'causal understanding' is defined and the inherent limits of human cognition. Given current technological and theoretical advancements, AI may already surpass human abilities in specific domains or possess the potential to do so. This ``surpassing'' manifests in AI’s ability to process complex data, uncover hidden patterns, and achieve greater efficiency in reasoning.

A key advantage of AI is its ability to process vast amounts of data and identify complex causal relationships, especially in fields where human sensory and cognitive abilities are limited. For example, AI can discern causal links in high-dimensional datasets that humans cannot intuitively comprehend. In genomics, AI analyzes millions of variables to identify causal pathways between genetic mutations and diseases, a task that could take human scientists decades to achieve \cite{whalen2022navigating}. In complex systems like climate models or economic systems, causality often exceeds human intuition \cite{athey2018impact,bracco2024machine}. AI employs techniques such as Bayesian networks or causal diagrams to uncover hidden causal structures in noisy data. 

AI also excels at efficiently validating and testing causal hypotheses on a scale far beyond human capabilities. AI can simulate counterfactual scenarios, such as ``What would happen if one variable changes?'' and test causal relationships computationally, whereas humans often rely on labor-intensive experiments or observations. For instance, AI can predict the effects of gene editing on diseases without conducting individual lab experiments \cite{gillmore2021crispr}. Similarly, AI can perform large-scale intervention simulations in virtual environments to evaluate the causal impacts of various variables. In drug development, AI can simulate molecular interactions with target proteins and predict optimal intervention strategies, significantly accelerating the causal reasoning process \cite{zhou2020artificial,mak2019artificial}.

Moreover, AI handles complex causal chains with greater depth and breadth than humans. In multilayered causal relationships—such as tracing the impact of molecular interactions on entire cellular networks—humans are often limited by the depth of chain they can analyze. AI, however, can construct and analyze deeply nested causal networks, from molecular behaviors to ecosystem-level dynamics. In nonlinear systems, such as ecology or economics, where causality involves feedback loops and dynamic interdependencies, AI uses dynamic modeling and simulation techniques to unravel these intricate mechanisms.

In summary, with its ability to process data on an unprecedented scale, identify hidden causal patterns, and analyze complex networks, AI has the potential to surpass human capabilities in efficiency and comprehensiveness when understanding causality.

\section{AI Technologies}

AI is rapidly evolving. LLMs based on the Transformer architecture \cite{vaswani2017attention}, such as GPT\footnote{\url{https://openai.com/chatgpt/overview/}} and Claude\footnote{\url{https://www.anthropic.com/claude}}, have been described by some researchers as ``sparks of general artificial intelligence'' \cite{bubeck2023sparks}. These models leverage powerful attention mechanisms and parallel processing capabilities, achieving significant advances in multimodal data processing, cross-domain knowledge integration, and step-by-step reasoning.

\subsection{Attention: Transformer Architecture}

The Transformer model is a deep learning architecture originally designed for machine translation tasks but now widely used across various natural language processing (NLP) applications \cite{vaswani2017attention, lin2022survey}. Its core feature is the attention mechanism, which allows the model to assign different weights to input elements based on their relevance when generating outputs, focusing on the most pertinent information. The Transformer is composed of multiple encoder and decoder layers, each utilizing self-attention mechanisms to capture long-range dependencies in sequences, enabling the model to understand contextual relationships between words in complex sentences.

In a Transformer, the attention mechanism is divided into ``self-attention'' and ``multi-head attention''. Self-attention enables each word in the input sequence to ``attend'' to all other words and calculate attention weights based on relevance, capturing inter-word relationships. Multi-head attention extends this by employing multiple parallel attention mechanisms to extract diverse semantic features, allowing the model to focus on different linguistic aspects simultaneously. This parallel processing and global focus make the Transformer superior to traditional networks for handling long texts and complex language structures.

Specifically, the self-attention mechanism generates a query vector $Q$, a key vector $K$, and a value vector $V$ for each word in the input sequence. By computing the dot product of $Q$ and $K$, the model quantifies the similarity between the current word (represented by $Q$) and all other words (represented by $K$). These similarities are transformed into attention weights

\begin{equation}
A(Q, K, V) = \text{softmax}\left(\frac{QK^T}{\sqrt{d_k}}\right)V.
\end{equation}

The $\text{softmax}$ function ensures that the attention weights $A$ form a probability distribution where the sum of each row equals 1. These attention weights $A$ are then multiplied by the value vectors $V$, generating a weighted output representation for the current word. This process enables the model to adjust each word's representation based on the global context, capturing long-range dependencies within the input sequence.

Multi-head attention extends self-attention by introducing multiple independent attention distributions for the same input sequence, allowing the model to capture diverse relationships or feature spaces. The multi-head attention is mathematically expressed as

\begin{equation}
\text{MultiHead}(Q, K, V) = \text{Concat}(\text{head}_1, \text{head}_2, \dots, \text{head}_h)W^O
\end{equation}

where each attention head $\text{head}_i$ is computed as

\begin{equation}
\text{head}_i = A(QW_i^Q, KW_i^K, VW_i^V)
\end{equation}

Here, $W_i^Q$, $W_i^K$, and $W_i^V$ are projection matrices for the query, key, and value vectors of each head, while $W^O$ is the linear transformation matrix for the final output. By introducing independent projection matrices for each head, the model can compute attention in different feature spaces, capturing diverse semantic information such as syntactic structures, word dependencies, and sentence semantics. Finally, the outputs of all attention heads are concatenated and linearly transformed into the final output.

The structure of the Transformer has intriguing parallels with concepts in physics. In physics, equations describe the behavior of objects within force fields, defining interaction strengths and directions through precise parameters. Similarly, the attention weight matrices in the Transformer numerically encode the relationships and directionalities between words. In physics, solutions to equations reveal the system's evolution and final states under certain initial conditions, analogous to the text generation process in Transformers, where the attention mechanism iteratively generates words to form complete sentences or paragraphs. The generated text can be seen as a ``solution'' to language under the constraints of input conditions and attention weight matrices.

Moreover, the superposition principle in quantum mechanics bears a resemblance to the Transformer's multi-head attention mechanism. In quantum mechanics, particles can exist in multiple superposed states until measurement collapses them into a single state. Similarly, the Transformer's multi-head attention mechanism allows each word to exist in multiple ``states'' or ``views'', capturing different linguistic features simultaneously. Only during final output generation are these states consolidated into a single word vector representation. This probabilistic nature mirrors how Transformers generate text—not as deterministic answers, but as outputs sampled from a probability distribution.

While the Transformer architecture focuses on producing linguistically coherent outputs, it does not inherently perform reasoning or understanding, which are central to physics. However, in large-scale, multi-layer neural networks, the nonlinear combination of many Transformer units may yield emergent properties resembling logic and causality. This phenomenon parallels the concept of ``emergence'' in complex systems theory, where high-level behaviors arise from the interactions of simpler units \cite{fromm2004emergence, floridi2020gpt,schaeffer2024emergent}, which also occurs in our neural system of brain. 

\subsection{Chain-of-Thought Reasoning}

In recent years, Chain-of-Thought (CoT) reasoning has emerged as a key innovation for enhancing the logical reasoning capabilities of LLMs \cite{wei2022chain,chu2023survey,feng2024towards}, it explicitly generates intermediate reasoning steps. Although LLMs excel at language generation and knowledge-based tasks, traditional approaches often struggle with logical consistency in complex problems. CoT reasoning bridges this gap by guiding the model to simulate human-like thought processes, enabling it to derive solutions step-by-step and construct complete logical chains from input to output. 

OpenAI's CoT reasoning lies in the introduction of ``Reasoning Tokens'' \footnote{\url{https://platform.openai.com/docs/guides/reasoning/how-reasoning-works}}. These tokens are specially designed internal representations that capture intermediate states and logical transitions within the model. Unlike traditional language tokens, which primarily represent words or sentence structures, reasoning tokens encode logical transitions and serve as connectors between successive reasoning steps. They are dynamically generated and stored in the model's context window, allowing the model to effectively utilize prior reasoning information and integrate it with the current task requirements.

Theoretically, CoT reasoning decomposes complex problems into a series of logically related sub-tasks. This process can be expressed mathematically as
\begin{equation}
y = f(x) = f_k \circ f_{k-1} \circ \cdots \circ f_1(x),
\end{equation}
where the input $x$ represents the problem, the output $y$ is the solution, and $f_i$ represents the model's operation at the $i$th reasoning step. Each operation function transforms the intermediate state $s_{i-1}$ into the next state $s_i$, ultimately completing the logical progression from the initial input $s_0 = x$ to the final answer $s_k = y$. This step-by-step decomposition not only simplifies the handling of complex problems but also makes the reasoning process more transparent and controllable.

Reasoning token generation relies on the attention mechanism within the Transformer architecture. At each reasoning step, the model computes the next intermediate state using the following formula
\begin{equation}
s_{i+1} = W^{(i)} \cdot A(s_i, C),
\end{equation}
where $W^{(i)}$ represents the weight matrix specific to the current step, and $A(s_i, C)$ captures the interaction between the current state $s_i$ and the context $C$. This mechanism allows the model to dynamically adjust during the reasoning chain, allocating computational resources based on problem complexity. The dynamic nature of reasoning tokens ensures logical consistency in multi-step reasoning tasks while avoiding information loss.

Training reasoning tokens involves a combination of supervised and reinforcement learning techniques to ensure that the generated reasoning chains are both logically sound and lead to correct final answers. During training, the model optimizes a composite loss function that considers both intermediate steps and final outcomes
\begin{equation}
\mathcal{L} = \lambda_1 \mathcal{L}_{\text{reasoning}} + \lambda_2 \mathcal{L}_{\text{output}},
\end{equation}
where $\mathcal{L}_{\text{reasoning}}$ measures the deviation of intermediate steps from the true logical sequence, and $\mathcal{L}_{\text{output}}$ focuses on the accuracy of the final answer. By tuning the weight coefficients $\lambda_1$ and $\lambda_2$, the model balances the quality of the reasoning chain and the correctness of the final output. Reinforcement learning \cite{sutton2018reinforcement} further enhances reasoning token training. The model optimizes the reasoning process using reward signals that evaluate both the logical accuracy of each step and the completeness of the reasoning chain
\begin{equation}
R = \sum_{i=1}^k \delta_i r_i + \delta_k r_k,
\end{equation}
where $r_i$ represents the reward value for the $i$th step, and $\delta_i$ indicates the importance weight of each step. This method ensures that the model focuses not only on the final answer but also on refining each step in the reasoning process, maintaining the coherence of the logical chain.

LLM models support long context windows, for example, OpenAI's o1-series models support context windows of up to 128k tokens \footnote{\url{https://platform.openai.com/docs/models}}. This enables the handling of highly complex tasks such as multi-step mathematical proofs, intricate program generation, and advanced scientific problem-solving. However, long context windows introduce the risk of information overload. To mitigate this, the model employs dynamic management mechanisms that prioritize storing key reasoning steps while compressing irrelevant information \cite{jiang2024long}. This resource optimization strategy ensures efficiency and accuracy in executing complex reasoning tasks.

Beyond improving accuracy, CoT significantly enhances model interpretability. By examining the generated reasoning chain, users can clearly understand how the model arrives at its answers. This transparency is especially valuable in fields like theoretical physics, medical diagnostics, and financial analysis, where high-confidence results are critical. Additionally, reasoning chains help users quickly identify errors, facilitating debugging and model improvement.

\subsection{Multimodal Processing}

Initially, LLMs were designed to focus on textual data \cite{zhao2023survey}. However, with advances in AI, LLMs have evolved to handle multimodal tasks\footnote{\url{https://openai.com/index/upgrading-the-moderation-api-with-our-new-multimodal-moderation-model/}}, including image generation, image-text alignment, speech processing, and multimodal question answering \cite{yin2024survey}. This transition is driven by the versatility of the Transformer architecture and the ability to pretrain on large-scale datasets. By incorporating multimodal capabilities, models like GPT can process and generate data across various modalities such as text, images, and video, enabling AI to develop a more comprehensive understanding of the objective world.

The implementation of multimodal processing in LLMs involves several key steps: representation, alignment, fusion, and reasoning. These steps form the technical foundation for enabling multimodal understanding, generation, and inference \cite{zhang2020multimodal,li2021align,li2024empowering}.

Modal Representation is the first step in multimodal processing. Each modality has its unique data structure. For example, textual data consists of discrete token sequences, while images are represented as 2D pixel matrices. LLMs use specialized feature extractors to map each modality's data into high-dimensional vector spaces. For textual data, embedding layers convert words or sentences into vectors. For visual data, pretrained vision models (such as ResNet\cite{he2016deep} or Vision Transformer, ViT\cite{dosovitskiy2020vit}) extract image features. This can be mathematically represented as
\begin{equation}
\mathbf{h}_t = f_{\text{text}}(x_{\text{text}}), \quad \mathbf{h}_v = f_{\text{vision}}(x_{\text{vision}}),
\end{equation}
where $x_{\text{text}}$ and $x_{\text{vision}}$ are the text and image inputs, $f_{\text{text}}$ and $f_{\text{vision}}$ are the corresponding feature extractors, and $\mathbf{h}_t$ and $\mathbf{h}_v$ are the high-dimensional semantic features.

Modal Alignment is a critical step in multimodal processing. Text and image modalities often differ significantly in data distribution and feature structures, necessitating a mapping into a shared semantic space for further fusion and reasoning. Alignment is typically achieved using projection matrices
\begin{equation}
\mathbf{z}_t = W_{\text{text}}\mathbf{h}_t, \quad \mathbf{z}_v = W_{\text{vision}}\mathbf{h}_v.
\end{equation}
Here, $W_{\text{text}}$ and $W_{\text{vision}}$ are learnable projection matrices that unify the features of each modality into the same dimensionality and semantic space. The aligned features $\mathbf{z}_t$ and $\mathbf{z}_v$ exhibit semantic consistency, enabling direct comparison and fusion across modalities.

Information Fusion is essential for cross-modal reasoning. In LLMs, fusion is typically achieved through the multi-head attention mechanism in the Transformer architecture. For cross-modal tasks, the model employs cross-attention to dynamically adjust the information weights between modalities
\begin{equation}
\text{CrossAttention}(Q, K, V) = \text{softmax}\left(\frac{QK^T}{\sqrt{d_k}}\right)V.
\end{equation}
In this mechanism, the query vector $Q$ may come from the text modality, while the key and value vectors $K$ and $V$ originate from the image modality. This allows the model to establish dynamic associations between modalities. For example, in scientific data analysis, the model can focus on relevant regions in an image based on the semantic features of a textual query, enabling precise cross-modal reasoning.



\section{Discussion and Conclusion: Technology for Understanding}

Building on the previous discussion of intuition and causality, we further explore how the Transformer architecture, Chain-of-Thought reasoning methods, and multimodal capabilities support and expand our understanding of these concepts. 

While these technologies mark significant advancements in artificial intelligence, it is important to first acknowledge that AI, particularly artificial general intelligence (AGI), is still in its infancy. Our ambitions are vast, substantial room for improvement remains.

The core of the Transformer architecture lies in the attention mechanism, enabling the model to globally attend to all elements in the input data and assign varying weights to them. This mechanism enables AI to process large amounts of information simultaneously, capturing complex relationships and forming what can be described as a ``synthetic intuition'' that goes beyond human intuition. The self-attention and multi-head attention mechanisms allow the model to focus on all parts of the input data without the limitations of human working memory. For instance, in natural language processing, the Transformer can understand the semantics and context of words based on their relationships within an entire sentence. This capability resembles an extended form of intuition, enabling AI to capture long-range dependencies and implicit relationships that humans might overlook.  Transformer models operate in high-dimensional vector spaces, capable of handling data with hundreds or even thousands of dimensions. This allows AI to discover patterns and structures in high-dimensional spaces, identifying relationships beyond human intuition. While the human brain struggles to intuitively imagine spaces beyond three dimensions, AI can compute and reason within high-dimensional spaces, uncovering intricate connections hidden in the data. This high-dimensional processing capability allows AI to identify key features and patterns in fields such as astrophysics, offering novel scientific insights.

The Chain-of-Thought reasoning method enables AI to solve complex problems step by step, decomposing them into logically connected sub-tasks. This approach helps AI navigate intricate causal chains, maintaining logical consistency and reducing errors that humans often face due to cognitive load and memory limitations. Reasoning tokens allow AI to represent and manipulate abstract reasoning concepts, systematically construct causal networks, test hypotheses, and evaluate counterfactuals. While humans handle causal relationships with limited chain depth, CoT enables AI to manage far longer and more complex causal chains. For instance, in climate modeling, economic systems, or galactic evolution, AI can explore intricate interactions and uncover causal relationships beyond human reach.

AI's multimodal capabilities allow it to process and integrate data from diverse sensory inputs, extending its range of ``perception'' and forming entirely new types of intuition. Through modal representation and alignment, AI maps data from different modalities—such as text, images, and audio—into a shared semantic space. This unified representation enables models to establish associations across modalities, allowing AI to perceive and understand complex scientific phenomena from multiple perspectives. For example, AI can simultaneously process telescope observation images, laboratory simulation data, and textual descriptions to comprehensively analyze the properties of celestial bodies. 

In addition, AI can process data beyond the range of human sensory perception, such as infrared and ultraviolet spectra, as well as gravitational wave and neutrino signals. By incorporating these data into models, AI can ``perceive'' physical phenomena that humans cannot directly experience, forming entirely new intuitions about the world. AI is capable of handling data across different spatial and temporal scales, from microscopic molecular structures to macroscopic cosmic formations, and from millisecond reactions to million-year evolutionary processes. This capability enables AI to uncover cross-scale patterns and causal relationships.

However, compared to the human brain, current AI technology appears relatively crude both in its mathematical formulation and practical performance. Moreover, AI still lacks many essential abilities. First, the human brain possesses a high degree of randomness, which grants humans creativity and flexibility, whereas existing machines lack true randomness. Second, humans can efficiently learn from single events and adapt rapidly to unknown environments, while AI models often require large amounts of data and repeated training to achieve similar performance. Furthermore, the training data for current models is primarily based on forward logic to answer well-defined questions, with limited coverage of complex backward reasoning or the diversity of causal relationships, this limitation makes it difficult for AI to correctly distinguish between necessary and sufficient conditions in reverse reasoning. More importantly, current AI lacks self-awareness and cannot comprehend the boundaries of its own knowledge, further limiting the authenticity and reliability of its intelligence.

In summary, while AI could potentially achieve and even surpass human-level understanding, current technologies remain in the early stages of development. AI is still like an infant in its cradle, requiring us to explore more data and develop new methods to nurture and educate it.



\end{document}